\pdfoutput=1 
\documentclass[11pt]{article}

\usepackage[font=libertinus, citestyle=numeric]{kurbanlab}
\usepackage{svg}

\DeclareAffiliation{hbku}{%
  College of Science and Engineering, Hamad Bin Khalifa University, Doha, Qatar}

\DeclareAffiliation{tamuq}{%
  Department of Computer and Electrical Engineering,
  Texas A\&M University at Qatar, Doha, Qatar}

\DeclareAffiliation{tamu}{%
  Department of Computer and Electrical Engineering,
  Texas A\&M University, College Station, TX, USA}

\DeclareAffiliation{iub}{%
  Luddy School of Informatics, Computing, and Engineering,
  Indiana University Bloomington, Bloomington, IN, USA}

\title{Waiting Buys Exactly Two Moments: Endogenous Identification in Prediction-Driven Allocation}
\Subtitle{What Waiting Reveals and What It Leaves Unidentified}
\RunningTitle{Waiting Buys Exactly Two Moments}

\Author{Rama AlHamidi}{tamuq}
\Author{Rasul Khanbayov}{hbku}
\Author{Erchin Serpedin}{tamu}
\Author[corresponding=hkurban@hbku.edu.qa, orcid=0000-0003-3142-2866]{Hasan Kurban}{hbku}

\CodeURL{https://github.com/KurbanIntelligenceLab/waiting-buys-two-moments}
\Venue{}          

\begin{document}
\maketitle

\begin{abstract}
A planner with a fixed budget of preventive help must decide whom to
help and when. Waiting reveals risk but removes intervention
opportunities as higher-risk individuals leave first. We characterize
this tradeoff under repeated noisy signals and risk-dependent attrition:
in the canonical model, a length-$\tau$ observation prefix identifies
exactly the latent risk moments
$m_j=\mathbb E[p^j]$ for $j=0,\ldots,2\tau-1$,
buying two new moments per additional period. These identify average next-period attrition among survivors but need not identify allocation welfare
or attrition among those prioritized for treatment.
Our observationally equivalent constructions establish unavoidable
timing regret at any sample size for a fixed observation length,
while exact certificates determine whether all compatible populations
share a unique optimal date. Matched experiments separate the effects
of estimation, bootstrap averaging, and regret weighting.
Longer observation often resolves timing ambiguity, but the optimal
date may already have passed. In our experiments, bootstrap averaging
and early stopping improve decisions, while regret weighting adds
no improvement in pooled mean squared regret: better statistical
fit need not yield better allocation welfare. The code and artifacts are available at \href{https://github.com/KurbanIntelligenceLab/waiting-buys-two-moments}{https://github.com/KurbanIntelligenceLab/waiting-buys-two-moments}
\end{abstract}

\printkeywords

\section{Introduction}
\label{sec:intro}

Prediction-driven allocation models, such as Novissi \citep{novissi}, HPU \citep{hpu}, and DEWS \citep{dews}, guide decisions about who should receive support when resources are insufficient to help everyone. For example, consider a hospital with more seriously ill patients than beds equipped to provide close monitoring and life-saving treatment. Doctors can use the information already available to decide who receives those beds, or wait for more test results to better identify who would benefit most. During that wait, however, some patients may become too ill to be saved. Choosing when to act therefore requires weighing the value of additional information against the opportunities to help that disappear during the delay.

\begin{figure}
    \centering
    \includegraphics[width=1\linewidth]{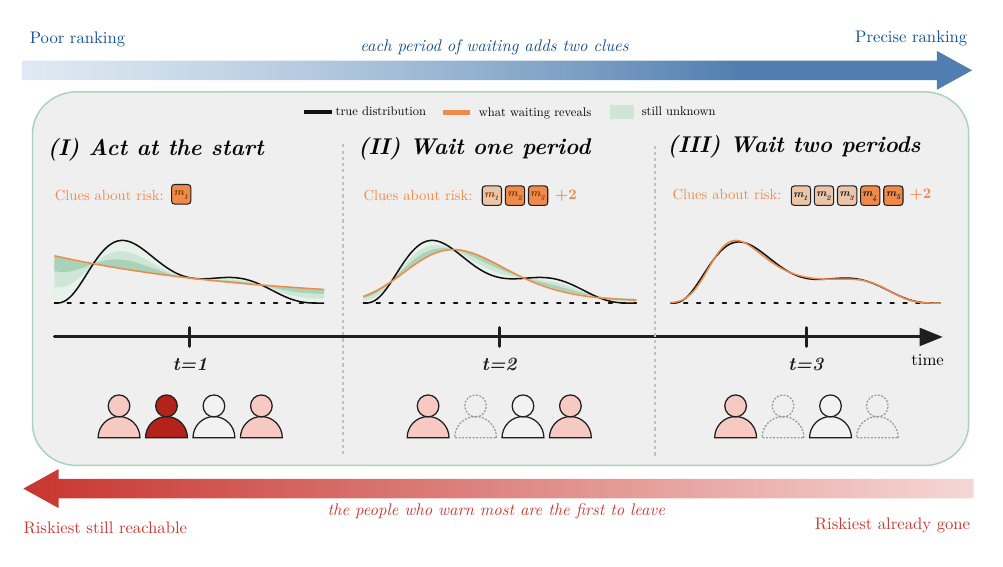}
    \caption{\textbf{Waiting buys exactly two moments, but the riskiest people leave first.} After t periods the planner observes exactly the first $2t - 1$ moments of the risk distribution (orange tiles), so its reconstruction (orange) of the true distribution (black) sharpens with every period of waiting (blue arrow). The people who generate those clues are the highest-risk ones, and they drop out first (dashed outlines): waiting improves the ranking of a population that has already lost the people most worth helping (red arrow).}
    \label{fig:teaser}
\end{figure}

Our review (Section~\ref{sec:related}) finds that existing timing results characterize when to intervene given the population’s risk distribution \citep{shirali2025hidden}. These results do not address whether a short observation period reveals enough about that distribution to determine whether acting now or waiting would lead to better outcomes.

This work closes this gap by establishing what a limited observation period reveals about the population’s risks and whether that information is enough to choose the best intervention date. We show that the observations can reveal how many people are expected to lose their opportunity for help during the next period and establish when these observations determine the best intervention date versus when they leave the decision unresolved. 

Furthermore, we find that uncertainty about the best time to intervene can persist even when the number of people observed becomes arbitrarily large. This raises a practical question: \textit{how do estimation, averaging, and decision criteria affect allocation performance on selected reference populations in this partially identified model?} To investigate this question, we compare risk estimation, bootstrap aggregation, and ambiguity-aware decision criteria using simulated populations with known welfare curves. Procedures recommend dates for a future population with the same risk distribution. Comparisons within a common information set isolate the effects of fitting, averaging, and selecting a date.

\begin{kilkey}
Waiting reveals average attrition among survivors, but not whether waiting is worthwhile; identical observations can imply different optimal allocation dates, even with arbitrarily large samples.
\end{kilkey}

Our contributions are:
\begin{enumerate}
    \item \textbf{Information and timing limits.}
    We characterize the moments revealed by observation,
    distinguish identified attrition from potentially
    unidentified welfare, and derive a regret floor
    that persists at every sample size.

    \item \textbf{Exact decision identification.} We reduce comparisons of nonlinear allocation welfare over an infinite-dimensional moment class to two-parameter branch-and-bound with polynomial certificates verified on $[0,1]$. This certifies a common optimal date or produces an exactly compatible counterexample, for both unrestricted and still-available dates.

    \item \textbf{Decision performance under limited information.}
    We compare risk estimation, bootstrap aggregation,
    and ambiguity-aware rules. Comparisons holding
    information fixed show that early stopping can
    outperform a converged likelihood fit, while
    regret weighting offers no improvement over
    ordinary averaging in the reported pooled results.
\end{enumerate}

\section{Related work}
\label{sec:related}

\paragraph{Identification and allocation timing}
Finite binomial mixtures identify finitely many mixing moments
\citep{wood1999binomial,lindsay1983geometry}, while heterogeneous
survival selects those remaining at risk
\citep{vaupel1979impact,elbers1982true,abbring2007unobserved}.
We characterize what joint signal-and-survival observations
identify and the implications for welfare and timing regret.
Adopting \citet{shirali2025hidden}'s model, we ask whether
limited observations identify the optimal date across
observationally consistent risk distributions; their analysis
takes the distribution as given. Their multi-date allocation
algorithm is not a directly comparable date-selection baseline:
our oracle is optimal only within single-date policies.
This complements work on predictive accuracy and allocation
design \citep{shirali2024allocation,perdomo2024relative,
fischer2025value,fischerabaigar2026metadesign}.

\paragraph{Decisions under partial identification}
Welfare and regret criteria under ambiguity are well studied
\citep{manski2022identification,christensen2022optimal,
montielolea2026decision,kitagawa2023mean}.
We examine how observation and attrition change ambiguity
and compare implemented procedures.
Existing optimality results require a separate derivation
here: \citet{montielolea2026decision} study Gaussian experiments;
\citet{kitagawa2023mean}, binary treatment choice with Gaussian
information and interval-identified effects.
Our setting uses multinomial signal-and-exit records,
allocation-date choices, nonlinear budget-constrained welfare,
and squared relative regret normalized by population-optimal
welfare. Our CMS bootstrap is therefore an approximate
comparison, not an established finite-sample optimal rule.
Unlike fixed-estimand sampling \citep{adusumilli2026sampling},
waiting removes treatment opportunities; unlike strategic
disclosure \citep{rosenthal2026identification}, information
restrictions arise from population dynamics.


\section{Problem Setup}
\label{sec:setup}

We adopt the model of \citet{shirali2025hidden}.
Individuals have fixed risks
$p_i\overset{\mathrm{iid}}{\sim}\mathcal P$ on $[0,1]$,
unknown to the planner. In each period, an active individual
produces a warning
$o_i^t\sim\mathrm{Bernoulli}(1-(1-p_i)^\gamma)$
and then fails with probability $p_i$ if untreated.
Conditional on risks, signal and failure draws are
independent across individuals and periods. Failure
permanently removes treatment eligibility. Our main
analysis uses $\gamma=1$.

The planner allocates treatment once before horizon $T$,
with budget $B$ covering a fraction $\beta=B/N$ of the
initial population. At date $t$, active individuals are
ranked by accumulated warnings
$y_i^t=\sum_{s=1}^{t}o_i^s$; treatment proceeds in descending
order, uniformly within ties. Treatment prevents failure
over the remaining $T-t$ opportunities, giving expected
benefit $u^t(p)=1-(1-p)^{T-t}$.
Let $W^t(\mathcal P;\beta)$ denote the resulting
mean-field welfare per initial individual. The earliest
optimal date is
\[
t^\star=\min\operatorname*{arg\,max}_{t\in\{1,\ldots,T-1\}}
W^t(\mathcal P;\beta).
\]
Appendix~\ref{app:welfare} gives the cohort masses, the
treated mass in each warning cohort, and the resulting welfare
functional explicitly.

We distinguish the observation length $\tau$ from the
candidate allocation date $t$. The theoretical prefix
ends after period-$\tau$ signals but before that period's
failures. It records earlier failures and warning counts
observed while individuals remain active. Cohort counts
and exit records restricted to this same window identify
the same population features
(Appendix~\ref{app:records}). Some finite-sample
implementations also use period-$\tau$ failures; these
information differences are stated in Experimental Design.


\section{Information and the Limits of Allocation Timing}
\label{sec:identification}

Throughout this section, $\gamma=1$ and identification
refers to exact population observation probabilities.
A quantity is identified if it is constant across all
risk distributions compatible with those probabilities.
Proofs appear in Appendix~\ref{app:proofs}.

\subsection{What waiting reveals.}
\begin{lemma}[Information accrual]
\label{lem:moments}
Let $\Pi_d$ denote polynomials of degree at most $d$.
The observation kernels
\[
p\mapsto\binom{s}{k}p^k(1-p)^{2s-1-k},
\qquad 1\le s\le\tau,\quad 0\le k\le s,
\]
span exactly $\Pi_{2\tau-1}$. Hence the population
observation probabilities determine moments through
degree $2\tau-1$ and depend only on them. The same holds
for exit records restricted to this observation window.
\end{lemma}

The first period identifies mean risk; each additional period
reveals two more moments through survivor signals.
Attrition removes treatment opportunities while changing
what signals reveal. Higher moments need not be identified,
though boundary moments can determine the distribution uniquely.
Appendix~\ref{app:gamma} considers other signal models.

\subsection{What the identified moments determine.}
\begin{proposition}[Identification asymmetry]
\label{prop:asym}
For a prefix $\tau\ge1$ with positive surviving mass,
the expected fraction lost during the next untreated
period is identified:
\[
\mathbb E^\tau[p]
=\frac{\mathbb E[p(1-p)^{\tau-1}]}
       {\mathbb E[(1-p)^{\tau-1}]}.
\]
Welfare at date $s$ involves kernels of degree $T+s-1$
and is identified whenever $s\le2\tau-T$.
\end{proposition}

The identified rate measures average survivor attrition,
not targeted delay costs: cohort attrition uses degree-$2\tau$
kernels and need not be identified
(Appendix~\ref{app:targeted-attrition}).
The sufficient, not necessary, welfare condition covers
no date unless $\tau\ge(T+1)/2$.

\subsection{Consequences for timing decisions.}
For the observed population moments
$m_j=\mathbb E_{\mathcal P}[p^j]$, define
\[
\mathcal M_\tau
=
\left\{
Q\text{ a probability distribution on }[0,1]:
\mathbb E_Q[p^j]=m_j,\quad 0\leq j\leq2\tau-1
\right\}.
\]
Uncertain welfare need not imply
an uncertain decision: compatible populations may share
an optimal date. When they do not, observational
equivalence imposes an unavoidable loss.

\begin{theorem}[Irreducible regret]
\label{thm:floor}
For $\mathcal Q_0,\mathcal Q_1\in\mathcal M_\tau$
with positive optimal welfare, define
\[
R_i(s)=1-\frac{W^s(\mathcal Q_i)}{\max_r W^r(\mathcal Q_i)},
\qquad
\zeta=\min_s\{R_0(s)+R_1(s)\},
\]
where dates range over $\{1,\ldots,T-1\}$.
Every rule based on a length-$\tau$ prefix, randomized
or not, incurs expected relative regret at least
$\zeta/2$ under at least one population, for every $N$.
\end{theorem}

Identical data laws imply identical recommendation distributions.
The bound is positive exactly when no optimal date is shared,
so larger samples cannot resolve ambiguity at a fixed prefix.

\section{Methodology}
\label{sec:methodology}

\subsection{Identification ambiguity}
\label{sec:identification-ambiguity}

For the reference population's unique optimal date $d$,
we bound each competitor's largest welfare advantage:
\[
V_\tau(d,r)=
\sup_{\mathcal P\in\mathcal M_\tau}
[W^r(\mathcal P)-W^d(\mathcal P)].
\]
Strictly negative upper bounds for all $r\ne d$ certify
that $d$ is uniquely optimal throughout $\mathcal M_\tau$.
A compatible population where a competitor outperforms
$d$ establishes ambiguity: $d$ fails there, while every
other date fails for the reference population.
We verify certificates and counterexamples exactly for both
unrestricted and still-available dates, leaving unresolved
cases undecided.

\subsection{Decision performance}
\label{sec:decision-performance}

Each procedure estimates or bounds welfare across dates
and recommends an intervention time.

\paragraph{Estimators and bootstrap methods}
We compare \textbf{certified nonparametric maximum likelihood
(NPMLE)} \citep{lindsay1983geometry,wood1999binomial},
a flexible likelihood benchmark;
\textbf{moment-matched Beta} \citep{shirali2025hidden}; and
\textbf{early-stopped expectation--maximization (EM)} \citep{dempster1977maximum}, testing whether further likelihood optimization improves decisions. All use
$\hat d\in\arg\max_d W^d(\hat{\mathcal{P}})$;
NPMLE and EM share information, while Beta uses a different observation
convention detailed below. \textbf{Bagged NPMLE} \citep{breiman1996bagging} averages bootstrap
welfare curves; \textbf{regret calibration} weights the same curves
by inverse optimal welfare, isolating regret weighting from averaging.

\paragraph{Ambiguity-aware rules}
We compare four implemented criteria. The \textbf{achieved-welfare LP}
selects the earliest candidate date whose approximate best-case achieved
welfare is not exceeded by any later date's approximate worst-case
welfare, with extrema taken over a moment confidence region.
The \textbf{worst-case envelope} instead compares against next-period
best-case full-information welfare, testing whether using achieved
welfare changes recommendations. Both scan dates from a fixed observation
prefix. \textbf{Maximin welfare} maximizes approximate worst-case achieved
welfare. \textbf{CMS bootstrap} adapts the bootstrap decision framework
of \citet{christensen2022optimal} to allocation timing,
minimizing bootstrap-averaged approximate worst-case regret
over distributions matching each draw's estimated moments,
relative to each population's optimal date. These finite-risk-grid extrema are not full-set
certificates, and refinement can change decisions
(Appendix~\ref{app:methods}).

\subsection{Evaluation \& Data analysis}
\label{sec:eval-analysis}
We evaluate recommendations using true population welfare:
$E=W^{\hat d}(\mathcal P;\beta)/
W^{t^\star}(\mathcal P;\beta)$ and $R=1-E$.
We report mean squared regret (MSR), the average of $R^2$,
and efficiency's tenth percentile ($p_{10}$);
lower MSR and higher $p_{10}$ are better.
Displayed MSR is scaled by $10^3$, not a percentage.
$N$ controls estimation sample size; evaluation uses
mean-field welfare, not realized finite-population allocations.

The main comparison retains 1,169 shared runs
(589 at $N=10^3$, 580 at $N=10^5$), excluding
31 with missing bagging records.
CMS returns all 1,200 decisions, including seven
using 19 rather than 20 bootstrap draws.
Inference uses paired bootstrap intervals and
within-family Holm correction
(Appendix~\ref{app:inference}).
Immediate action means date 1 for a future population
or date $\tau$ when restricted to remaining dates.

\section{Experimental Design}
\label{sec:experiments}

\subsection{Identification ambiguity}

\begin{figure*}
    \centering
    \includegraphics[width=1\linewidth]{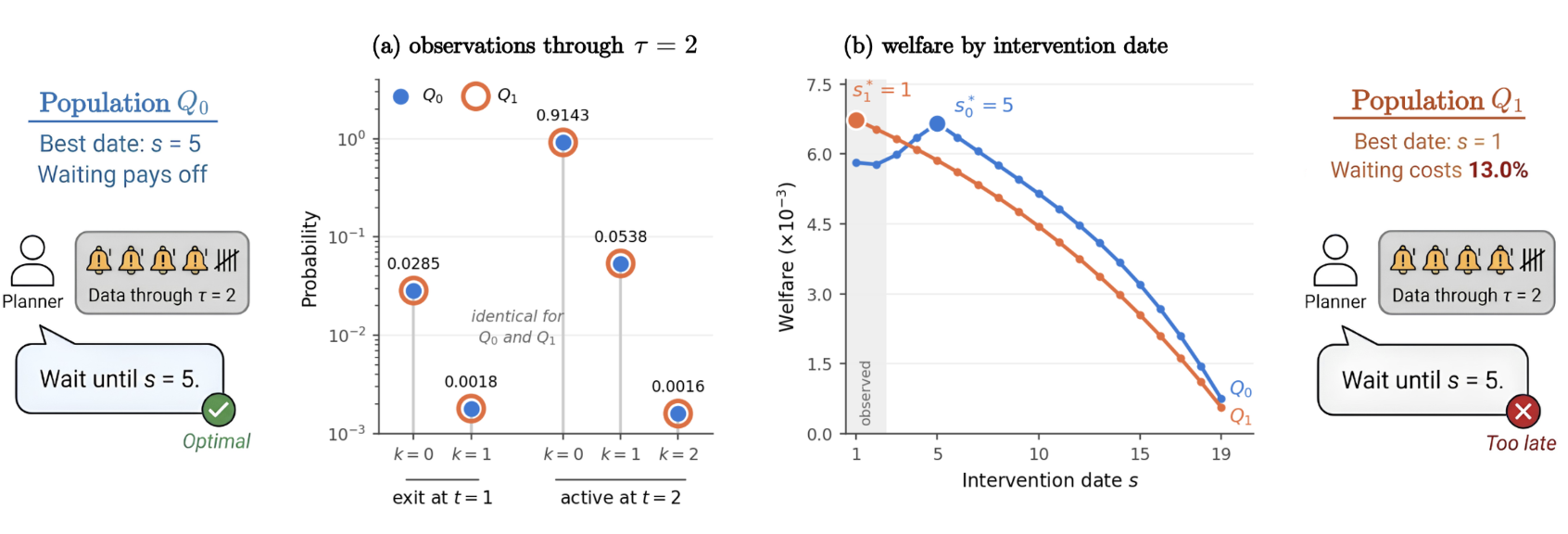}
    \caption{\textbf{Identical observations, different optimal dates.} The two populations have matching observation probabilities through $\tau=2$ (a), but different welfare curves (b): $Q_0$ is optimized at $s=5$, whereas $Q_1$ is optimized at $s=1$. The illustrated recommendation $s=5$ is optimal for $Q_0$ but incurs relative welfare loss under $Q_1$. Dates are evaluated as policies for a future population with the same risk distribution.}
    \label{fig:indistinguishable-pair}
\end{figure*}

We use 30 configurations formed by crossing five risk distributions---Beta$(1,32)$, Beta$(1,8)$, Beta$(1,4)$, a mixture containing a high-risk subgroup, and a bimodal distribution---with horizons $T\in\{12,20\}$ and budget ratios $\beta\in\{0.0005,0.002,0.01\}$. We fix $\gamma=1$ and examine observation lengths $\tau\in\{2,3,4\}$, giving 90 configuration--prefix pairs. Each analysis uses exact population moments through degree $2\tau-1$, removing sampling error so that any remaining uncertainty reflects identification ambiguity.

\paragraph{Indistinguishable populations}
For each configuration and prefix, we numerically
minimize and maximize horizon survival,
$\mathbb E_{\mathcal Q}[(1-p)^{T-1}]$, over a
moment-matching set on a 2,001-point risk grid.
We reconstruct rational weights on the selected
supports and verify exact compatibility for all
90 pairs. We then compare their welfare curves
and full optimal-date sets and compute the
regret lower bound from Theorem~\ref{thm:floor}
in rational arithmetic. These pairs exhibit
ambiguity but are not selected to maximize
the regret floor.

\paragraph{Optimal-date identification}
For each configuration and prefix, we apply the
certification procedure to all dates $\{1,\ldots,T-1\}$
and separately to available dates $\{\tau,\ldots,T-1\}$.
Certificates cover the full risk interval $[0,1]$ and
are independently verified, together with counterexamples,
in exact arithmetic. Among identified available dates,
we distinguish immediate action from further waiting.
Appendix~\ref{app:certification-protocol} details
computational budgets.

\subsection{Decision performance}

We use the same 30 population configurations at $\gamma=1$, with sample sizes $N\in\{10^3,5\times10^3,2\times10^4,10^5\}$ and observation lengths $\tau\in\{2,3,4\}$. Each setting has 20 independent trials, giving 600 planned runs per sample-size--prefix combination. Procedures recommend an allocation date for a future population with the same risk distribution, and recommendations are evaluated using its true welfare curve.

\paragraph{Information supplied to each procedure} Beta, the two worst-case rules, maximin, and CMS use cohort-based moment estimates through order $2\tau-1$, corresponding to the information in the identification experiment. NPMLE, early-stopped EM, bagging, and regret calibration instead use exit records including period-$\tau$ failures, revealing one further moment at the population level. Cross-family comparisons therefore reflect differences in both information and procedure. CMS recomputes cohort moments within each bootstrap draw and uses an exact moment-matching set, projecting infeasible estimates when necessary. Maximin and the worst-case rules instead use a covariance-based confidence region.

\paragraph{Estimation and bootstrap comparisons}
We compare certified NPMLE, moment-matched Beta, and EM
stopped after 30, 55, or 80 iterations. Bagging and regret calibration share the same retained ensemble from 200 attempted bootstrap fits.

\paragraph{Ambiguity-aware comparisons} We evaluate the worst-case envelope, achieved-welfare LP, and maximin welfare procedures as numerical approximations, reporting sensitivity to refinement separately from their decision performance. The CMS comparison uses 20 bootstrap draws at $\tau=2$ and $N\in\{10^3,10^5\}$, with approximate worst-case regret calculations. Its matched comparison retains 1,169 runs. The estimator comparison at these same sample sizes and prefix retains 1,122 runs across Beta, certified NPMLE, and all three EM stopping counts. Failure handling follows the conventions stated in the evaluation subsection.


\section{Results}

\subsection{Identification ambiguity}

Table~\ref{tab:identification-ambiguity} shows that populations
with identical observation laws can differ substantially in
both welfare and preferred intervention dates. This uncertainty
cannot be removed by observing more individuals over the same
prefix. Where the constructed pairs have no shared optimal
date, Theorem~\ref{thm:floor} establishes a welfare cost that
every observation-based rule must incur under at least one
population. Longer observation reduces the exhibited welfare
spreads and regret floors, but the two quantities describe
different consequences: uncertainty about intervention value
need not imply a costly timing mistake. Figure~\ref{fig:indistinguishable-pair} illustrates the disagreement; a date optimal for one compatible
population can incur substantial regret under another.

The full-set analysis in Table~\ref{tab:date-identification}
shows that additional observations identify the optimal
date in more configurations. Identifying that date
does not require identifying the entire risk distribution. It also reveals
ambiguity missed by the selected extreme pairs: at $\tau=3$,
the full-set search finds more ambiguous configurations than
the pair construction exhibits. Agreement between two compatible
populations therefore cannot establish that a decision is
identified. The remaining ambiguous cases at the longest
prefix reflect insufficient population information, rather
than sampling error or an unresolved computation.

In a majority of the identified configuration--prefix cases,
the best overall date precedes the observation endpoint.
Such knowledge can guide allocation in a future population,
but cannot recover a missed opportunity in the observed one.
The available-date panel of Table~\ref{tab:date-identification}
addresses the remaining decision directly. Restricting the
options can resolve disagreement about past dates; among
the resulting identified decisions, acting now is much more
common than waiting further. Thus, on this benchmark,
additional information often settles when to stop waiting,
while some populations remain indistinguishable even about
the best action still available.

\begin{table}[t]
\centering
\small
\caption{\textbf{Identification ambiguity across observation lengths.}
Selected population pairs match moments through
degree $2\tau-1$ and therefore generate identical
prefix observation laws. Compatibility, welfare
trajectories, and regret separation are verified
in exact rational arithmetic for all 90 pairs.
Welfare spread is the maximum relative difference
between the two welfare curves across candidate dates.
Regret bounds are exhibited lower bounds, not
necessarily sharp over the compatible set.
Median regret bounds are calculated among pairs
with disjoint optimal-date sets.}
\label{tab:identification-ambiguity}

\renewcommand{\arraystretch}{1.15}
\begin{tabularx}{\linewidth}{@{\hspace{1em}}Xrrr@{}}
\toprule
& \multicolumn{3}{c}{\textbf{Observation length}} \\
\cmidrule(lr){2-4}
\multicolumn{1}{@{}l}{\textbf{Measure}}
    & $\tau=2$ & $\tau=3$ & $\tau=4$ \\
\midrule
\multicolumn{4}{@{}l}{\textit{Welfare uncertainty}} \\
Median exhibited welfare spread
    & $42.1\%$ & $27.3\%$ & $15.2\%$ \\
Interquartile range
    & $27.8$--$63.8\%$ & $14.0$--$38.8\%$ & $2.7$--$27.4\%$ \\
Largest exhibited welfare spread
    & $85.8\%$ & $74.3\%$ & $50.5\%$ \\
\addlinespace
\multicolumn{4}{@{}l}{\textit{Timing disagreement}} \\
Configurations with disjoint optimal-date sets
    & $29/30$ & $11/30$ & $3/30$ \\
\addlinespace
\multicolumn{4}{@{}l}{\textit{Regret separation $\zeta/2$}} \\
Median bound among disagreement cases
    & $1.37\%$ & $0.080\%$ & $0.0091\%$ \\
Largest exhibited bound
    & $6.34\%$ & $1.35\%$ & $0.47\%$ \\
\bottomrule
\end{tabularx}
\end{table}

\begin{table}[t]
\centering
\small
\caption{\textbf{Do the observations determine the best intervention date?}
Counts across 30 configurations per observation length.
Identified means one date is uniquely optimal for every
compatible population; ambiguous means no date is optimal
for all of them. All cases were resolved after the extended
pass. Indented rows partition the identified available-date cases.}
\label{tab:date-identification}
\renewcommand{\arraystretch}{1.15}
\begin{tabularx}{\linewidth}{@{}Xccc@{}}
\toprule
\textbf{Outcome} & $\tau=2$ & $\tau=3$ & $\tau=4$ \\
\midrule
\multicolumn{4}{@{}l}{\textit{All dates: $1,\ldots,T-1$}} \\
\quad Best date identified
    & $\mathbf{1/30}$ & $\mathbf{16/30}$ & $\mathbf{27/30}$ \\
\quad Best date ambiguous
    & $29/30$ & $14/30$ & $3/30$ \\
\addlinespace
\multicolumn{4}{@{}l}{\textit{Available dates: $\tau,\ldots,T-1$}} \\
\quad Best available date identified
    & $\mathbf{3/30}$ & $\mathbf{18/30}$ & $\mathbf{28/30}$ \\
\qquad Act now ($s=\tau$)
    & $3/30$ & $14/30$ & $27/30$ \\
\qquad Wait ($s>\tau$)
    & $0/30$ & $4/30$ & $1/30$ \\
\quad Best available date ambiguous
    & $27/30$ & $12/30$ & $2/30$ \\
\bottomrule
\end{tabularx}
\end{table}

\subsection{Decision performance}

The main benchmark evaluates reference populations.
A supplementary experiment evaluates recommendations
under both members of 30 observationally equivalent
pairs, using a common pre-failure information set
(Appendix~\ref{app:matched-pairs}).
Using the same observations, likelihood, risk grid,
and initialization, EM stopped after 55 iterations
has lower pooled worst-member mean regret than
certified NPMLE at both sample sizes.
This comparison covers selected compatible pairs,
not the full moment class.

Table~\ref{tab:decision-performance} shows that early-stopped
EM achieves the lowest MSR and highest lower-tail welfare
efficiency among the tested non-oracle procedures at both
sample sizes. CMS improves on certified NPMLE and, at the
larger sample size, outperforms both bagging and regret
calibration. Its relative performance therefore depends
on sample size. These results concern the implemented
20-draw CMS approximation and do not establish the
performance of an exactly optimized or bootstrap-converged
version.

The comparison across all three stopping counts
appears in Appendix, Table~\ref{tab:estimator-comparison}.
EM and certified NPMLE share observations and a
welfare-maximizing decision rule, showing that
further likelihood optimization need not improve
allocation decisions on this benchmark. A supplementary study finds lower pooled MSR than
NPMLE at 55 iterations for each of eight tested
initializations. This does not establish a general
early-stopping advantage or robustness to
observation-model changes. In Table~\ref{tab:decision-performance},
both bootstrap procedures improve on a single
NPMLE fit, while ordinary bagging achieves lower
MSR than regret calibration at both sample sizes.
These results support averaging without demonstrating
an additional benefit from regret weighting.

The approximate ambiguity-aware procedures yield mixed
results (Appendix, Table~\ref{tab:pooled-rule-comparison}).
The envelope procedure matches immediate action on all
reported metrics. Maximin welfare improves mean and
median efficiency and optimal-date agreement, but
offers little improvement in lower-tail efficiency.
These numerical results are sensitive to scan refinement
and are distinct from the exact identification certificates.

A supplementary trajectory study at $\tau=2$ supports
a population-dependent explanation for early stopping.
EM55 has a positive median error in the estimated
third moment, $m_3=\mathbb E[p^3]$, across all eight
tested initializations. Continued fitting often lowers
this estimate and shifts recommendations toward earlier
intervention, which is costly on the reference populations.
At $N=10^3$, regret increases from EM55 to the
million-update endpoint in 660 of 1,500 datasets.
In post-hoc interventions, retaining EM55's $m_3$
while adopting the endpoint's $m_1$ and $m_2$
preserves the EM55 recommendation in 470 of 609
feasible cases (77.2\%) and removes 89.3\% of their
aggregate additional regret. Counting the 51 infeasible
cases as unrepaired reduces the latter figure to 77.4\%.
These results identify changes in $m_3$ as a major
channel of deterioration and support a beneficial
bias rather than uniformly better risk estimation.
Holding $m_3$ also blocks many improvements, and
the early-stopping advantage can reverse under
compatible alternative populations.

\begin{table*}[t]
\centering
\caption{\textbf{Overall decision performance.}
Results at $\tau=2$ on 1,169 matched runs across the two
sample sizes. MSR is mean squared relative regret multiplied
by $10^3$; MSR gain is its reduction relative to certified
NPMLE. $p_{10}$ is the 10th percentile of welfare efficiency.
Early-stopped EM uses 55 iterations.}
\label{tab:decision-performance}
\small
\setlength{\tabcolsep}{7pt}
\begin{tabular}{@{}lrrrrrr@{}}
\toprule
& \multicolumn{3}{c}{$N=10^3$}
& \multicolumn{3}{c}{$N=10^5$} \\
\cmidrule(lr){2-4}\cmidrule(lr){5-7}
\textbf{Procedure}
& MSR $\downarrow$ & MSR gain $\uparrow$ & $p_{10}$ $\uparrow$
& MSR $\downarrow$ & MSR gain $\uparrow$ & $p_{10}$ $\uparrow$ \\
\midrule

\multicolumn{7}{@{}l}{\textit{Reference policies}} \\
\quad Oracle
& 0.00 & +4.61 & 1.0000
& 0.00 & +1.60 & 1.0000 \\
\quad Act immediately
& 8.64 & $-4.02$ & 0.8535
& 8.85 & $-7.26$ & 0.8201 \\

\addlinespace
\multicolumn{7}{@{}l}{\textit{Single fit and bootstrap aggregation}} \\
\quad Certified NPMLE
& 4.61 & +0.00 & 0.8672
& 1.60 & +0.00 & 0.9658 \\
\quad Bagged NPMLE
& 3.49 & +1.13 & 0.8683
& 0.59 & +1.01 & 0.9878 \\
\quad Regret-calibrated
& 3.95 & +0.67 & 0.8681
& 0.63 & +0.96 & 0.9878 \\

\addlinespace
\multicolumn{7}{@{}l}{\textit{Ambiguity-aware decision rule}} \\
\quad CMS bootstrap (20 draws)
& 3.86 & +0.76 & 0.8683
& 0.35 & +1.24 & 0.9921 \\

\addlinespace
\multicolumn{7}{@{}l}{\textit{Early stopping}} \\
\quad Early-stopped EM (55 iterations)
& \textbf{0.21} & \textbf{+4.40} & \textbf{0.9964}
& \textbf{0.00} & \textbf{+1.59} & \textbf{0.9969} \\
\bottomrule
\end{tabular}

\medskip
\begin{minipage}{0.98\textwidth}
\footnotesize
\textit{Notes.}
Bold indicates the best reported non-oracle point
estimate. CMS uses the pre-failure prefix; the
likelihood-based methods additionally observe
period-$\tau$ failures. Cross-family comparisons
therefore change both information and procedure.
\end{minipage}
\end{table*}

\section{Conclusion}
\label{sec:conclusion}

In the canonical model, each additional observation
period identifies two new risk moments. These determine
average attrition among survivors, but need not identify
targeted attrition, welfare, or the optimal intervention
date. Exact certificates distinguish decisions shared
by all compatible populations from persistent ambiguity;
longer observation can resolve timing only after the
best date has passed.

On the tested reference populations, early stopping
improves decisions and ordinary bagging achieves lower
pooled MSR than regret weighting. A supplementary
comparison with aligned information also finds lower
pooled worst-member regret for early-stopped EM than
certified NPMLE over 30 constructed compatible pairs.
Post-hoc moment interventions support the explanation
that early stopping retains an upward bias in the
estimated third moment, limiting costly shifts toward
earlier intervention on the reference populations.
This bias helps decisions without necessarily improving
risk estimation; its benefit can reverse under
compatible alternatives. The results do not establish robustness over the full moment class; numerical approximation and limited trial counts remain limitations.

\begin{kilrepro}
Section~\ref{sec:setup} specifies the model and observation
window, and Section~\ref{sec:methodology} defines the
decision procedures. Appendix~\ref{app:proofs} provides
proofs; Appendix~\ref{app:identification-computation}
describes exact pair reconstruction and full-set
certification; and Appendix~\ref{app:methods} documents
estimation, initialization, bootstrap filtering,
numerical optimization, and inference. Observation-window
differences across procedures are stated explicitly.

The supplementary material provides frozen configurations
and seeds, simulation and estimation code, per-run outputs,
and reproduction scripts. Exact certificates and
counterexamples are checked by a separate rational-arithmetic
verifier. Numerical decision procedures are reported with
their approximation settings, retained sample sizes, and
known limitations. The implementation generating the
reported regret-calibration results is identified explicitly
to distinguish it from the separate randomized procedure
in the codebase.
\end{kilrepro}

\bibliography{references}

\appendix
\section{Model details and proofs}
\label{app:proofs}

\subsection{Welfare and allocation}
\label{app:welfare}

For $\gamma=1$, define the cohort mass and welfare contribution
\[
q_k^t(\mathcal P)
=
\int \ell_{t,k}(p)\,d\mathcal P(p),
\qquad
v_k^t(\mathcal P)
=
\int u^t(p)\ell_{t,k}(p)\,d\mathcal P(p),
\]
where
\[
\ell_{t,k}(p)
=
\binom{t}{k}p^k(1-p)^{2t-k-1},
\qquad
u^t(p)=1-(1-p)^{T-t}.
\]
The kernel $\ell_{t,k}(p)$ is the joint probability,
conditional on risk $p$, of surviving the first $t-1$
failure opportunities and accumulating $k$ warnings
through period $t$. With budget fraction $\beta$, the treated mass in cohort $k$ is
\[
a_k^t
=
\min\left\{
q_k^t,\,
\left(\beta-\sum_{j=k+1}^{t}q_j^t\right)_+
\right\}.
\]
Thus,
\[
W^t(\mathcal P;\beta)
=
\sum_{k:q_k^t>0}
a_k^t\frac{v_k^t}{q_k^t}.
\]
This implements descending priority by warning count and
uniform treatment within ties.

\subsection{Observation window and information accrual}
\label{app:records}

The theoretical prefix ends after period-$\tau$ signals
and before period-$\tau$ failures. Its exit records have
kernels
\[
e_{s,k}(p)=p\,\ell_{s,k}(p),
\quad s<\tau,
\qquad
r_{\tau,k}(p)=\ell_{\tau,k}(p).
\]
These record earlier failures and the warning counts of
individuals still eligible at the final observation.

\paragraph{Proof of Lemma~\ref{lem:moments}}
The date-$s$ cohort kernels are $(1-p)^{s-1}$ times
a Bernstein basis of $\Pi_s$. Their combined span is
therefore
\[
V_\tau=\sum_{s=1}^{\tau}(1-p)^{s-1}\Pi_s
\subseteq\Pi_{2\tau-1}.
\]
At $\tau=1$, $V_1=\Pi_1$. If
$V_{\tau-1}=\Pi_{2\tau-3}$, the polynomials
$(1-p)^{\tau-1}p^{\tau-1}$ and
$(1-p)^{\tau-1}p^\tau$ supply independent directions
of degrees $2\tau-2$ and $2\tau-1$.
Hence $V_\tau=\Pi_{2\tau-1}$ by induction.

Every exit-record kernel also has degree at most
$2\tau-1$. Conversely, exit-record probabilities
recover earlier cohort probabilities: conditional on
a terminal warning count, warning locations are
exchangeable, with a hypergeometric distribution
independent of $p$. Summing over terminal records
therefore recovers the cohort probabilities through
a distribution-independent linear transformation.
Both representations consequently span
$\Pi_{2\tau-1}$.
\hfill$\square$

This equivalence requires the same observation window.
The implemented NPMLE family additionally observes
period-$\tau$ failures, introducing degree-$2\tau$
kernels. Its information therefore differs from the
cohort-based moment procedures and the population
identification experiment.

\subsection{Identification asymmetry}

\paragraph{Proof of Proposition~\ref{prop:asym}}
Let $S_\tau=\mathbb E[(1-p)^{\tau-1}]>0$.
The fraction of survivors lost in the next untreated
period is
\[
1-\frac{S_{\tau+1}}{S_\tau}
=
\frac{\mathbb E[p(1-p)^{\tau-1}]}
     {\mathbb E[(1-p)^{\tau-1}]}.
\]
The numerator and denominator have degrees $\tau$
and $\tau-1$, so both are identified by the prefix.

The cohort kernel $\ell_{t,k}$ has degree at most
$2t-1$, and $u^t\ell_{t,k}$ has degree at most
$T+t-1$. Consequently, all cohort masses and welfare
contributions, and hence the allocation fractions
and welfare, are identified whenever
$T+t-1\leq2\tau-1$, or $t\leq2\tau-T$.
This is sufficient, not necessary: welfare can be
constant over a compatible set without every input
being separately identified.
\hfill$\square$

\paragraph{Attrition in targeted cohorts}
\label{app:targeted-attrition}
For a positive-mass warning cohort $k$, next-period
attrition is
\[
h_{\tau,k}(\mathcal Q)
=
\frac{\mathbb E_{\mathcal Q}[p\,\ell_{\tau,k}(p)]}
     {\mathbb E_{\mathcal Q}[\ell_{\tau,k}(p)]}.
\]
The denominator is identified, but the numerator
has degree $2\tau$, beyond the identified moment
span. Attrition within warning cohorts, including
those prioritized for treatment, therefore need
not be identified. Identifying average attrition
among survivors does not establish that the
welfare consequences of waiting are known.

\subsection{Irreducible timing regret}
\label{app:regret}

\paragraph{Proof of Theorem~\ref{thm:floor}}
Identical individual record laws imply identical joint
laws for every number $N$ of independent records.
Any observation-based rule therefore has the same
recommendation distribution, say $\pi$, under
$\mathcal Q_0$ and $\mathcal Q_1$. Using the theorem's
relative regrets,
\[
\sum_{i=0}^{1}
\mathbb E_{\mathcal Q_i}[R_i(\hat d)]
=
\sum_d\pi(d)\{R_0(d)+R_1(d)\}
\geq\zeta.
\]
At least one expected regret is at least $\zeta/2$.
Independent randomization can be included in the
rule's input, so the argument also covers randomized
rules.
\hfill$\square$

Because the candidate set is finite and regrets are
nonnegative, $\zeta>0$ exactly when the populations
share no optimal date. Different tie-broken dates
alone do not establish this condition.

\paragraph{Available-date version}
For any common nonempty candidate set
$S\subseteq\{1,\ldots,T-1\}$, define
\[
R_{i,S}(s)
=
1-\frac{W^s(Q_i)}{\max_{r\in S}W^r(Q_i)},
\qquad
\zeta_S
=
\min_{s\in S}\{R_{0,S}(s)+R_{1,S}(s)\},
\]
assuming positive denominators. The same proof
gives expected relative regret at least
$\zeta_S/2$ under one population for every rule
taking values in $S$. Choosing
$S=\{\tau,\ldots,T-1\}$ gives the available-date
result, normalized by each population's best
remaining date.

Treatment effectiveness is assumed, not estimated:
identifying untreated attrition does not identify
causal treatment benefits.

\subsection{Other signal models}
\label{app:gamma}

For signal probability $1-(1-p)^\gamma$, write
$q=1-p$. The date-$s$ kernels are $q^{s-1}$ times
a Bernstein basis in $q^\gamma$, giving observable span
\[
V_\tau^\gamma
=
\operatorname{span}
\{q^{\gamma m+s-1}:1\leq s\leq\tau,\ 0\leq m\leq s\}.
\]
Distinct real powers are linearly independent on $(0,1)$,
so dimension counts distinct exponents: $0,\ldots,2\tau-1$
at $\gamma=1$, and $\tau(\tau+3)/2$ distinct $(s,m)$
exponents for irrational $\gamma$. The two-moment rate
is therefore specific to the canonical model; all
experiments use $\gamma=1$. Time-varying risks or conditionally dependent signals and failures can change this span and require a separate identification analysis.

\section{Identification computations}
\label{app:identification-computation}

\subsection{Extreme-pair construction}

For each population moment vector, we minimize and
maximize $\mathbb E_{\mathcal Q}[(1-p)^{T-1}]$
over distributions supported on a 2,001-point grid
in $[0,1]$, subject to
\[
w_g\geq0,\qquad
\sum_g w_g=1,\qquad
\sum_gw_gp_g^j=m_j,\quad 1\leq j\leq2\tau-1.
\]
We evaluate both resulting welfare curves and compute
their regret separation. These selected pairs exhibit
ambiguity; they need not maximize welfare spread or
the regret floor over the compatible set. Failure
to find disagreement is not a certificate of
identification.

For a selected pair $(Q_0,Q_1)$, the reported
welfare spread is
\[
\Delta_W
=
\max_{1\leq s\leq T-1}
\frac{|W^s(Q_0)-W^s(Q_1)|}
{\max\{W^s(Q_0),W^s(Q_1)\}},
\]
with a zero contribution when both welfare values
are zero. Table~\ref{tab:identification-ambiguity}
summarizes this quantity across configurations.

For all 90 configuration--prefix pairs at
$\tau\in\{2,3,4\}$, we reconstruct nonnegative
rational weights on numerically selected supports.
We verify unit mass and moment matching through
degree $2\tau-1$ exactly, then evaluate welfare
trajectories and regret separation in rational
arithmetic. Full optimal-date sets are checked
using exact equality; every reconstructed
population has a unique optimal date.
The two populations' optimal-date sets are
disjoint in 29, 11, and 3 configurations at
$\tau=2,3,4$, respectively.

\subsection{Certification criterion and protocol}
\label{app:certification-protocol}

Within each candidate set, let $d$ be the reference
population's unique optimal date. Strictly negative
upper bounds on $V_\tau(d,r)$ for every $r\ne d$
certify unique optimality throughout the compatible
set. A compatible population with
$W^r>W^d$ refutes universal optimality: $d$ fails
for that population, while every other date fails
for the reference population.

The implementation uses branch-and-bound with exact
Bernstein-polynomial nonnegativity checks and rational
counterexamples. Certificates cover the full risk
interval $[0,1]$; certificates and counterexamples
are independently verified in exact arithmetic.

Each reference-date comparison allows up to 3,000
boxes or 900 seconds. Unresolved comparisons receive
an extended pass allowing up to 30,000 boxes or two
hours. Both unrestricted and available-date analyses
use this protocol. Cases unresolved after extension
remain undecided.

\paragraph{Exact moment class}
Set $n=2\tau-1$ and define
\[
\psi_j(p)=p^j(1-p)^{n-j},
\qquad
\mu_j=\mathbb E_{\mathcal P}[\psi_j(p)],
\quad j=0,\ldots,n.
\]
These are exact population moments, not sample
estimates. The compatible set is
\[
\mathcal M_\tau
=
\left\{
Q\text{ a probability measure on }[0,1]:
\mathbb E_Q[\psi_j]=\mu_j,\ j=0,\ldots,n
\right\}.
\]
The mixed polynomials form a basis for polynomials
of degree at most $n$, so these constraints are
equivalent to matching raw moments through order
$n$. In particular,
\[
\sum_{j=0}^n\binom{n}{j}\psi_j(p)=1.
\]

\paragraph{Primal and dual welfare representations}
Using the cohort kernels from Appendix~\ref{app:welfare},
write
\[
q_{s,k}(Q)=\mathbb E_Q[\ell_{s,k}],
\qquad
v_{s,k}(Q)=\mathbb E_Q[\ell_{s,k}u^s].
\]
With $f_k$ denoting treatment fractions, welfare is
\[
W^s(Q)
=
\max_{\substack{0\leq f_k\leq1\\
\sum_kf_kq_{s,k}(Q)\leq\beta}}
\sum_kf_kv_{s,k}(Q).
\]
For positive-mass cohorts, increasing the warning
count multiplies the conditional risk likelihood
by a positive constant times $p/(1-p)$.
The conditional risk distributions are therefore
ordered by monotone likelihood ratio. Since
$u^s(p)$ is nondecreasing, the value densities
$v_{s,k}/q_{s,k}$ are nondecreasing in $k$.
The fractional-knapsack optimum treats cohorts
in descending warning order. A zero-mass cohort
also has zero utility mass and contributes
nothing to either objective or budget.

An optimal allocation consequently belongs to
\[
f_{s,k}(\theta)
=
\min\{1,\max\{0,\theta-(s-k)\}\},
\qquad \theta\in[0,s+1],
\]
giving
\[
W^s(Q)
=
\max_{\substack{\theta\in[0,s+1]\\
f_s(\theta)\cdot q_s(Q)\leq\beta}}
f_s(\theta)\cdot v_s(Q).
\]
Linear-programming duality also gives
\[
W^s(Q)=\min_{\lambda\in[0,1]}g_s(\lambda,Q),
\]
\[
g_s(\lambda,Q)
=
\lambda\beta+
\sum_{k=0}^s
\bigl(v_{s,k}(Q)-\lambda q_{s,k}(Q)\bigr)_+.
\]
Restricting the dual price to $[0,1]$ loses no
optimum because $0\leq v_{s,k}\leq q_{s,k}$.

\paragraph{Pairwise reduction}
Combining the primal representation at date $r$
with the dual representation at date $d$ yields
\[
V_\tau(d,r)
=
\sup_{\substack{
Q\in\mathcal M_\tau,\ \theta\in[0,r+1],\
\lambda\in[0,1]\\
f_r(\theta)\cdot q_r(Q)\leq\beta}}
G(Q,\theta,\lambda),
\]
where
\[
G(Q,\theta,\lambda)
=
f_r(\theta)\cdot v_r(Q)-\lambda\beta
-\sum_{k=0}^d
\bigl(v_{d,k}(Q)-\lambda q_{d,k}(Q)\bigr)_+.
\]
The optimization over $\lambda$ is a supremum
because the dual minimum is subtracted.

\paragraph{A bound for each parameter box}
For
$B=[\theta_l,\theta_u]\times[\lambda_l,\lambda_u]$,
define
\[
A_B(p)
=
\sum_{k=0}^r
f_{r,k}(\theta_u)\ell_{r,k}(p)u^r(p),
\]
\[
C_B(p)
=
\sum_{k=0}^r
f_{r,k}(\theta_l)\ell_{r,k}(p),
\qquad
D_{B,k}(p)
=
\ell_{d,k}(p)\bigl(u^d(p)-\lambda_u\bigr).
\]
For every feasible point in this box,
$\mathbb E_Q[C_B]\leq\beta$ and
\[
G(Q,\theta,\lambda)
\leq
\mathbb E_Q[A_B]-\lambda_l\beta
-\sum_{k=0}^d
\bigl(\mathbb E_Q[D_{B,k}]\bigr)_+.
\]

Choose $y\in\mathbb R^{n+1}$, $\eta\geq0$, and
$\omega\in[0,1]^{d+1}$. Define
\[
R_B(p)
=
\sum_{j=0}^n y_j\psi_j(p)
+\eta C_B(p)
+\sum_{k=0}^d\omega_kD_{B,k}(p)-A_B(p).
\]
If $R_B(p)\geq\ell_B$ throughout $[0,1]$,
integration against any compatible $Q$, together
with $\omega_kx\leq x_+$, gives
\[
G(Q,\theta,\lambda)
\leq
U_{\mathrm{cert}}(B)
:=
\sum_{j=0}^n y_j\mu_j
+\eta\beta-\lambda_l\beta
+\max\{0,-\ell_B\}.
\]
Thus $U_{\mathrm{cert}}(B)$ bounds every feasible
population and parameter pair in the box.

\paragraph{Exact polynomial checks and coverage}
A polynomial is bounded below by its smallest
Bernstein coefficient on an interval. Exact
subdivision refines this bound while preserving
validity. We use this construction to obtain
$\ell_B$ over the entire risk interval $[0,1]$.
Moments, multipliers, endpoints, and polynomial
coefficients are checked in rational arithmetic.
Floating-point optimization proposes multipliers;
it does not determine certificate validity.

The search partitions $[0,r+1]\times[0,1]$
into unit-width strips and recursively splits
unresolved boxes. A leaf is accepted only when
$U_{\mathrm{cert}}(B)<0$. A finite covering by
accepted leaves therefore proves $V_\tau(d,r)<0$.
The implementation additionally checks that
all boxes lie inside the domain and that their
exact total area is $r+1$. Coverage rests on
the recursive partition; area alone would not
exclude gaps and overlaps in an arbitrary
collection of boxes.

\paragraph{Exact compatible witnesses}
Let $A=\sum_iw_i\delta_{x_i}$ be a rational atomic
probability measure and choose rational
$\varepsilon\in(0,1)$. Seek
\[
Q=(1-\varepsilon)A+\varepsilon B,
\qquad
dB=(1+g)\,d\mathcal P,
\]
where $g$ has degree at most $n$.
For the exact residuals
\[
e_j=\mathbb E_A[\psi_j]-\mu_j,
\]
solve
\[
\mathbb E_{\mathcal P}[\psi_jg]
=
-\frac{1-\varepsilon}{\varepsilon}e_j,
\qquad j=0,\ldots,n.
\]
Direct substitution gives
$\mathbb E_Q[\psi_j]=\mu_j$ exactly.
Since $A$ and $\mathcal P$ have unit mass,
the identity
$\sum_j\binom{n}{j}\psi_j=1$ implies
$\mathbb E_{\mathcal P}[g]=0$.
Hence $B$ and $Q$ also have unit mass.

An exact Bernstein check of $1+g\geq0$ on $[0,1]$
is sufficient for nonnegativity. It is not a
necessary condition because positivity outside
the support of $\mathcal P$ is unnecessary.
Welfare is then evaluated by exact rational
greedy allocation. A witness is accepted only
if $W^r(Q)>W^d(Q)$ strictly, establishing
$V_\tau(d,r)>0$.

\paragraph{Scope of resolution}
The strict certificate and witness tests apply to
both unrestricted and available-date sets.
Comparisons with $V_\tau(d,r)=0$, or those exhausting
the computational budget without either outcome,
remain undecided.

\section{Finite-sample implementation}
\label{app:methods}

\subsection{Configurations and observation conventions}
\label{app:benchmark}

The five risk distributions are
$\mathrm{Beta}(1,32)$, $\mathrm{Beta}(1,8)$,
$\mathrm{Beta}(1,4)$,
$0.95\,\mathrm{Beta}(1,32)+0.05\,\mathrm{Beta}(8,2)$,
and
$0.5\,\mathrm{Beta}(3,57)+0.5\,\mathrm{Beta}(12,28)$.
Crossing these with $T\in\{12,20\}$ and
$\beta\in\{0.0005,0.002,0.01\}$ gives 30 configurations.

The full finite-sample grid uses
$N\in\{10^3,5\times10^3,2\times10^4,10^5\}$,
$\tau\in\{2,3,4\}$, and 20 trials per setting:
7,200 planned runs. Procedures share simulated
populations and records within each trial, but their
observation conventions differ as described in
Appendix~\ref{app:records}. Recommendations are
evaluated for a future population with the same
risk distribution. Ties are resolved toward the
earliest date.

\subsection{Estimation and bootstrap aggregation}

The likelihood estimators fit nonnegative mixing
weights on a fixed 200-point risk grid. Certified
NPMLE checks the directional-derivative optimality
condition on that grid with tolerance $10^{-8}$.
This numerical likelihood certificate is distinct
from a full-set optimal-date certificate.
Early-stopped EM uses the same likelihood framework
with 30, 55, or 80 iterations; the main comparison
uses 55.

Bagging and regret calibration use the same fitted
bootstrap ensemble, with 200 attempted draws per
retained run. The primary fit and bootstrap fits
start from uniform weights on the 200-point risk
grid; neither reported procedure uses a warm start.
Robustness to initialization is not evaluated.

Let $J_\tau$ denote the number of individual-record
cells in the likelihood model. A fit is rejected
if its support contains more than $J_\tau$ points
or its log likelihood is non-finite. An invalid
primary fit causes the run to be skipped; an invalid
bootstrap fit is removed from both procedures.
Let $\mathcal B_+$ index the bootstrap fits retained
by these checks. For each retained fit, define
\[
\hat W_b^t=W^t(\hat P_b;\beta),
\qquad
M_b=\max_{1\leq u\leq T-1}\hat W_b^u.
\]
When all retained $M_b$ are finite and strictly
positive, the criteria are
\[
\hat t_{\mathrm{bag}}
\in\arg\max_{1\leq t\leq T-1}
\sum_{b\in\mathcal B_+}\hat W_b^t,
\qquad
\hat t_{\mathrm{RC}}
\in\arg\max_{1\leq t\leq T-1}
\sum_{b\in\mathcal B_+}\frac{\hat W_b^t}{M_b}.
\]
Ties are resolved toward the earliest date.
Regret calibration changes the weighting of welfare
curves while holding the fitted ensemble fixed.

If any retained $M_b$ is non-finite or non-positive,
regret calibration returns no valid decision rather
than discarding that replicate individually.
An empty retained ensemble also yields no decision.
Neither condition occurs in the 7,199 archived
ensembles.

Of 7,200 planned runs, one is skipped because of
an invalid primary fit. Among the remaining runs,
6,792 retain all 200 draws and 407 retain between
194 and 199. Overall, 519 of 1,439,800 attempted
bootstrap fits are discarded (0.036\%). The archive
preserves retained counts but not the individual
discard reasons, so the breakdown between excess
support size and non-finite likelihood is unavailable.

\paragraph{Reproduction}
The reported bagging and regret-calibration results are
generated by
\path{experiments/scripts/bagged_full_b200.py},
which evaluates both rules on a shared bootstrap ensemble.
The regret-calibration implementation is
\texttt{expected\_regret\_rule}; the separately implemented
\texttt{regret\_calibrated()} function uses a different,
randomized sequential procedure and does not generate
the reported row. A targeted reproduction is provided by
\texttt{experiments/scripts/repro\_reported\_regret\_calibrated.py}.

\subsection{Approximate ambiguity-aware procedures}

On a risk grid, write
$\mathcal P_w=\sum_g w_g\delta_{p_g}$.
Both $q_k^t(w)$ and $v_k^t(w)$ are linear in $w$,
but $W^t(\mathcal P_w;\beta)$ generally is not.
For a boundary cohort $k^\star$ with $q_{k^\star}^t(w)>0$
and
\[
\sum_{j>k^\star}q_j^t(w)
\leq\beta
\leq
\sum_{j\geq k^\star}q_j^t(w),
\]
the welfare is
\[
W^t(\mathcal P_w;\beta)
=
\sum_{j>k^\star}v_j^t(w)
+
\left(\beta-\sum_{j>k^\star}q_j^t(w)\right)
\frac{v_{k^\star}^t(w)}{q_{k^\star}^t(w)}.
\]
If the total surviving mass is at most $\beta$, all
survivors are treated and welfare is $\sum_j v_j^t(w)$.
Thus, restricting the distribution to a finite grid does
not itself yield a linear welfare objective.

\paragraph{Numerical welfare extrema}
For candidate date $s$, define treatment fractions
\[
f_{s,k}(\theta)
=
\min\{1,\max\{0,\theta-(s-k)\}\},
\qquad \theta\in[0,s+1].
\]
These fractions prioritize higher-warning cohorts.
The treated masses $f_{s,k}(\theta)q_k^s(w)$ still
depend on the candidate distribution.

At each scanned $\theta$, the supremum calculation
solves
\[
\max_{w\in\hat{\mathcal C}_\tau}
\sum_{k=0}^s f_{s,k}(\theta)v_k^s(w)
\quad\text{subject to}\quad
\sum_{k=0}^s f_{s,k}(\theta)q_k^s(w)\leq\beta.
\]
This is a linear program in $w$, and its allocation
can underfill the budget. Its objective is no
greater than the optimal allocation welfare at
the returned distribution. Maximizing over a
finite collection of thresholds therefore
underestimates the grid-restricted supremum
in exact arithmetic.

Fractional-knapsack duality gives
\[
W^s(\mathcal P_w;\beta)
=
\min_{\lambda\in[0,1]}
\left[
\lambda\beta+
\sum_{k=0}^s
\bigl(v_k^s(w)-\lambda q_k^s(w)\bigr)_+
\right].
\]
At each scanned $\lambda$, the infimum calculation
minimizes this expression over
$w\in\hat{\mathcal C}_\tau$, using nonnegative
auxiliary variables for the positive parts.
Restricting $\lambda$ to finitely many values
overestimates the grid-restricted infimum in
exact arithmetic. Both approximation directions
refer to the finite support grid. HiGHS \citep{huangfu2017parallelizing} solves
the individual linear programs numerically,
introducing additional floating-point error.

Default scans use 21 allocation thresholds and
41 dual prices, followed by three local refinement
rounds with nine points per round. Fine scans use
201 initial points for each variable and seven
refinement rounds with 41 points per round.
Both use the same 201-point risk support grid.

Retaining previous evaluations makes a running
maximum nondecreasing and a running minimum
nonincreasing. Independently configured scans,
however, need not evaluate nested sets of points.
We therefore combine default and fine outputs
explicitly:
\[
\tilde U_c(s)
=
\max\{\tilde U_{\mathrm{default}}(s),
       \tilde U_{\mathrm{fine}}(s)\},
\]
\[
\tilde L_c(s)
=
\min\{\tilde L_{\mathrm{default}}(s),
       \tilde L_{\mathrm{fine}}(s)\}.
\]
Here $\tilde U$ approximates the supremum and
$\tilde L$ approximates the infimum; neither
is a certified bound.

An underestimated supremum and an overestimated
infimum can create apparent dominance. Errors in
approximate infima can also change the maximin
ranking. The resulting recommendations are
therefore numerical heuristics, and their
refinement sensitivity is reported separately
from exact identification certificates.

The worst-case rules and maximin use the linear
cohort-moment
estimator and its multinomial covariance. Empirical path-cell
frequencies are transformed linearly into the moment vector
$\hat b$, and the covariance is transformed by the same
linear map. Let $n_c$ count individuals in mutually exclusive path
cells, with $\sum_c n_c=N$, and let $\hat f_c=n_c/N$.
If $A_\tau$ is the fixed path-cell-to-moment map, then
\[
\hat b=A_\tau\hat f,
\qquad
\hat\Sigma_b
=
\frac{1}{N}A_\tau
\left[
\operatorname{diag}(\hat f)
-\hat f\hat f^\top
\right]A_\tau^\top.
\]
Here $\hat\Sigma_b$ estimates the covariance of
$\hat b$, rather than the covariance of a single
individual's contribution. The constant moment $m_0=1$
is imposed exactly and excluded from standardization.
Let $d=2\tau-1$, let $\hat b_+$ contain the
nonconstant moments, and let $C$ be their estimated
covariance matrix. Write
\[
C=V\operatorname{diag}(\lambda_1,\ldots,\lambda_d)V^\top.
\]
The implementation constructs
\[
L=\operatorname{diag}(h_1,\ldots,h_d)V^\top,
\qquad
h_i=
\begin{cases}
\lambda_i^{-1/2},
& \lambda_i>10^{-10}\max\{\lambda_{\max},10^{-300}\},\\
0,&\text{otherwise}.
\end{cases}
\]
Thus, $L^\top L$ is a thresholded pseudoinverse of $C$.
With $H_{jg}=p_g^j$ for $j=1,\ldots,d$, the region is
\[
\hat{\mathcal C}_\tau
=
\left\{
w\geq0:\ \mathbf 1^\top w=1,\quad
\left\|L(Hw-\hat b_+)\right\|_\infty
\leq c_{\alpha,d}
\right\},
\]
where
\[
c_{\alpha,d}
=
\Phi^{-1}\left(
\frac{1+(1-\alpha)^{1/d}}{2}
\right).
\]
We use $\alpha=0.05$ and a 201-point uniform risk grid.
The critical value uses $d=2\tau-1$ even when eigenvalues
are truncated. Truncated directions receive no direct
constraint from the whitened box. The constant moment
is enforced by $\mathbf 1^\top w=1$.
This is a nominal Gaussian-approximation construction;
exact finite-sample coverage is not established.

CMS uses cohort-based moments through order three
at $\tau=2$. Each of its 20 bootstrap draws
multinomially resamples the empirical path-cell counts
and recomputes the moments and covariance.
The inner optimization uses an exact moment-matching
set. If that set is empty, the moment estimate undergoes
covariance-weighted projection onto the feasible cone
before the set is reconstructed. Projection and
infeasibility indicators are recorded separately.
CMS then selects the date minimizing the average
approximate worst-case regret across draws.
Neither bootstrap convergence nor exact inner
optimization is established for this implementation.

\subsection{Evaluation and inference}
\label{app:inference}

Matched comparisons retain only runs returning a
decision for every included procedure. Exclusions
and retained sample sizes are reported separately
for each comparison.

Inference uses 2,000 configuration-cluster bootstrap
replicates (seed 20260906), resampling the 30 configurations
with their trials. Separately for each metric and
$(N,\tau)$ cell, Holm correction covers six comparisons
of regret calibration against certified NPMLE, NNLS,
parametric Beta, achieved-welfare LP, worst-case envelope,
and immediate action. None of the 12 lower-tail comparisons
against certified NPMLE is significant after correction;
this does not establish equivalence. Adjustment does not
extend across metrics or cells, and these tests do not
establish significance for early-stopping or CMS comparisons.

\section{Supporting decision-performance results}
\label{app:additional-results}

\subsection{Estimator choice and early stopping}

Table~\ref{tab:estimator-comparison} uses 525 matched
runs at $N=10^3$ and 597 at $N=10^5$.
Beta's nonpositive estimated variance excludes 75
and three runs, respectively, from every row.
This sample differs from
Table~\ref{tab:decision-performance}.
Beta uses cohort moments, whereas NPMLE and EM
use richer exit records; only the latter comparison
holds information fixed.

\begin{table}[t]
\centering
\small
\caption{\textbf{Estimator choice at $\tau=2$.}
MSR is multiplied by $10^3$; $p_{10}$ is the tenth
percentile of welfare efficiency. Rounded zeros
do not imply zero regret.}
\label{tab:estimator-comparison}
\begin{tabular}{@{}lrrrr@{}}
\toprule
& \multicolumn{2}{c}{$N=10^3$}
& \multicolumn{2}{c}{$N=10^5$} \\
\cmidrule(lr){2-3}\cmidrule(lr){4-5}
\textbf{Estimator} & MSR & $p_{10}$ & MSR & $p_{10}$ \\
\midrule
Certified NPMLE
& 3.37 & 0.8933 & 1.55 & 0.9876 \\
Parametric Beta
& 4.03 & 0.8536 & 3.81 & 0.8536 \\
EM, 30 iterations
& 0.19 & 0.9964 & 0.05 & 0.9964 \\
EM, 55 iterations
& 0.23 & 0.9964 & 0.00 & 0.9969 \\
EM, 80 iterations
& 0.27 & 0.9964 & 0.00 & 0.9969 \\
\bottomrule
\end{tabular}
\end{table}

Across all 12 $(N,\tau)$ settings, each stopping
count reduces MSR relative to certified NPMLE.
The 30-iteration estimate improves $p_{10}$ in
11 settings; the 55- and 80-iteration estimates
improve it in all 12. These are point-estimate
comparisons, not additional hypothesis tests.

\subsection{Approximate ambiguity-aware criteria}

Table~\ref{tab:pooled-rule-comparison} pools all
30 configurations, four sample sizes, three
observation lengths, and 20 trials per setting.
Of 7,200 planned runs, 11 have empty estimated
sets; all four rows use the remaining 7,189 runs.
The reported values combine the default and
fine scans as described in Appendix~\ref{app:methods}.

\begin{table}[t]
\centering
\small
\caption{\textbf{Approximate ambiguity-aware procedures.}
$p_{10}$, median, and mean summarize welfare efficiency;
MSR is multiplied by $10^3$. Exact is agreement with
the reference optimal date. All rows use the same
7,189 runs with nonempty estimated sets, out of
7,200 planned runs.}
\label{tab:pooled-rule-comparison}
\begin{tabular}{@{}lrrrrr@{}}
\toprule
\textbf{Rule} & $p_{10}$ & Median & Mean & MSR & Exact (\%) \\
\midrule
Act immediately
& 0.8535 & 0.9475 & 0.9251 & 8.693 & 6.66 \\
Worst-case envelope
& 0.8535 & 0.9475 & 0.9251 & 8.693 & 6.66 \\
Achieved-welfare LP
& 0.8535 & 0.9539 & 0.9299 & 8.032 & 8.43 \\
Maximin welfare
& 0.8536 & 0.9933 & 0.9604 & 5.006 & 32.35 \\
\bottomrule
\end{tabular}
\end{table}

The envelope procedure matches immediate action
on every reported metric, while achieved-welfare
LP offers a modest improvement. Maximin welfare
improves mean and median efficiency and selects
the reference optimal date more often, but its
tenth-percentile efficiency barely changes.
The gains therefore do not extend comparably
to the measured lower tail.

Refinement changes 324 of 801 achieved-welfare-LP
firing decisions (40.4\%) and 22 of 7,189 evaluated
maximin decisions (0.3\%). The denominators differ.
These changes demonstrate sensitivity to numerical
approximation, but do not by themselves determine
which recommendation is correct. They are separate
from the exact identification verification.

\subsection{Matched-information evaluation on compatible populations}
\label{app:matched-pairs}

We evaluate ten non-CMS procedures on both members
of the 30 constructed $\tau=2$ pairs underlying
Table~\ref{tab:identification-ambiguity}, including
the pair with zero regret separation.
Pair reconstruction uses exact rational arithmetic
(Appendix~\ref{app:identification-computation}).

\paragraph{Populations and observations}
Each pair induces identical probabilities for every
observable path through period-2 warnings, before
period-2 failures. The aggregated record cells are
$e_{1,0},e_{1,1},r_{2,0},r_{2,1},r_{2,2}$
(Appendix~\ref{app:records}).
For each pair, sample size, and seed, all procedures
receive one dataset drawn from this shared law;
none receives period-2 failures or the true population.
Each selected date is scored against both welfare
curves. Dates range over $\{1,\ldots,T-1\}$,
representing policies for a future population.

\paragraph{Procedures and computational settings}
Table~\ref{tab:matched-pair-risk} lists the ten
procedures; CMS is excluded.
NPMLE and EM share the common-window likelihood,
a uniform 200-point grid on $[0,1]$, and uniform
initial weights. The NPMLE optimality tolerance
is $10^{-8}$.
Bagging and regret calibration share 40 attempted
bootstrap fits, versus 200 in the main benchmark.
Moment-based procedures use estimates from the
same observation window.
The envelope, achieved-welfare LP, and maximin
implementations use a 101-point grid, nine allocation
thresholds, eleven dual prices, and one five-point
refinement round. The main benchmark instead uses
201 grid points, 21 thresholds, 41 dual prices,
and three nine-point refinement rounds.
These are numerical approximations, not exact
identification certificates.

\paragraph{Trials and matching}
For each pair and $N\in\{10^3,10^5\}$, we use seeds
$\{1001,\ldots,1005\}$, giving 300 planned datasets.
Joint validity across all ten procedures retains
137/150 trials at $N=10^3$ and 143/150 at $N=10^5$,
with three to five trials per pair.
NPMLE returns invalid fits on nine datasets;
Beta fails because of nonpositive estimated variance
on eleven other datasets; the remaining procedures always return
decisions. Results therefore condition on joint
validity. Restoring all planned trials and assigning
date 1 on failure changes pooled pair risk by at
most 0.0021 across procedures and sample sizes.

\paragraph{Risk and uncertainty}
For pair $j$, let $S_{j,N}$ be its common retained trial
set and let $\hat d_{j,N,s}^{\,a}$ be the date selected
by procedure $a$ on trial $s$. Using the relative regret
$R_{j,i}$ under member $i\in\{0,1\}$, define
\[
\hat\rho_{j,N}^{\,a}
=
\max_{i\in\{0,1\}}
\frac{1}{|S_{j,N}|}
\sum_{s\in S_{j,N}}
R_{j,i}\!\left(\hat d_{j,N,s}^{\,a}\right),
\qquad
\hat\rho_N^{\,a}
=
\frac{1}{30}\sum_{j=1}^{30}\hat\rho_{j,N}^{\,a}.
\]

The maximum follows trial averaging within each
population and precedes averaging across pairs;
this differs from mean trialwise maximum regret
and from MSR.

We use 2,000 paired bootstrap replicates, resampling retained trials within
each pair jointly across procedures and recomputing
the maximum before pooling.
The 95\% percentile intervals are exploratory,
unadjusted for multiplicity, and conditional on the
fixed pairs and retained samples. Only three to
five trials per pair support these intervals.

\begin{table}[t]
\centering
\small
\caption{\textbf{Worst-member mean regret over 30 compatible pairs.}
Procedures share pre-failure information and matched
trials. Values are relative-regret fractions;
bold marks the lowest point estimate per column.}
\label{tab:matched-pair-risk}
\begin{tabular}{@{}lrr@{}}
\toprule
Procedure & $N=10^3$ & $N=10^5$ \\
\midrule
Certified NPMLE       & 0.0648 & 0.0387 \\
EM, 30 iterations     & 0.0375 & 0.0383 \\
EM, 55 iterations     & \textbf{0.0357} & 0.0339 \\
EM, 80 iterations     & 0.0373 & 0.0341 \\
Bagged NPMLE          & 0.0558 & \textbf{0.0302} \\
Regret calibration    & 0.0576 & \textbf{0.0302} \\
Moment-matched Beta   & 0.0655 & 0.0670 \\
Achieved-welfare LP   & 0.0627 & 0.0408 \\
Worst-case envelope   & 0.1171 & 0.1167 \\
Maximin welfare       & 0.1121 & 0.0508 \\
\bottomrule
\end{tabular}
\end{table}

\begin{table}[t]
\centering
\small
\setlength{\tabcolsep}{4pt}
\caption{\textbf{Paired differences from certified NPMLE}
in pooled worst-member mean regret.
Negative values favor the listed procedure.
Brackets show exploratory, unadjusted 95\%
percentile-bootstrap intervals conditional on
the fixed pairs and retained trials.}
\label{tab:matched-pair-differences}
\begin{tabular}{@{}lcc@{}}
\toprule
Procedure & $N=10^3$ & $N=10^5$ \\
\midrule
EM, 55 iterations
& $-0.0292\;[-0.0362,-0.0242]$
& $-0.0048\;[-0.0109,-0.0022]$ \\
\addlinespace
Bagged NPMLE
& $-0.0090\;[-0.0153,-0.0032]$
& $-0.0085\;[-0.0140,-0.0047]$ \\
\addlinespace
Regret calibration
& $-0.0072\;[-0.0141,-0.0009]$
& $-0.0085\;[-0.0140,-0.0047]$ \\
\bottomrule
\end{tabular}
\end{table}

\paragraph{Results and interpretation}
EM55 has the lowest pooled pair risk at $N=10^3$;
bagging and regret calibration tie at $N=10^5$
(Table~\ref{tab:matched-pair-risk}).
EM55's paired differences from NPMLE are negative
at both sample sizes, with exploratory intervals
excluding zero
(Table~\ref{tab:matched-pair-differences}).
These results do not imply improvement on every
pair or both members, establish worst-case performance
over the full moment class, or explain why early
stopping helps.

\FloatBarrier

\end{document}